\documentclass[a4paper,fleqn]{cas-dc}

\usepackage[authoryear]{natbib}

\usepackage[dvipsnames,svgnames,table]{xcolor}
\usepackage{color, colortbl}

\usepackage{todonotes}
\usepackage{caption}
\usepackage{subcaption}
\usepackage{comment}
\usepackage{amsmath}

\usepackage{tabularx, ragged2e, array, arydshln,adjustbox, booktabs, tablefootnote, multirow, makecell}

\usepackage{soul}
\usepackage{changepage,threeparttable} 

\begin{document}
\let\WriteBookmarks\relax
\def\floatpagepagefraction{1}
\def\textpagefraction{.001}
\shorttitle{Rethinking Medical Knowledge Aware LLMs}
\shortauthors{Domingo-Aldama et~al.}

\title [mode = title]{To Adapt or not to Adapt, Rethinking the Value of Medical Knowledge-Aware Large Language Models}

\title[mode = title]{To Adapt or not to Adapt, Rethinking the Value of Medical Knowledge-Aware Large Language Models}

\cortext[cor1]{Corresponding author}
\fntext[equal]{These authors contributed equally to this work.}

\author[1]{Ane {G. Domingo-Aldama}¹*}[orcid=0009-0000-8202-1099]
\ead{ane.garciad@ehu.eus}

\author[1]{Iker {De La Iglesia}¹}[orcid=0000-0002-4141-992X]
\ead{iker.delaiglesia@ehu.eus}

\author[1]{Maitane {Urruela}}[orcid=0009-0007-4463-6325]
\ead{maitane.urruela@ehu.eus}

\author[1]{Aitziber {Atutxa}}[orcid=0000-0003-4512-8633]
\ead{aitziber.atucha@ehu.eus}

\author[1]{Ander {Barrena}}[orcid=0000-0003-2024-0362]
\ead{ander.barrena@ehu.eus}

\affiliation[1]{organization={University of the Basque Country}, 
                city={Bilbao},
                state={Biscay},
                country={Spain}}

\begin{abstract} 
BACKGROUND: Recent studies have shown that domain-adapted large language models (LLMs) do not consistently outperform general-purpose counterparts on standard medical benchmarks, raising questions about the need for specialized clinical adaptation. \\

\noindent METHODS: We systematically compare general and clinical LLMs on a diverse set of multiple choice clinical question answering tasks in English and Spanish. We introduce a perturbation based evaluation benchmark that probes model robustness, instruction following, and sensitivity to adversarial variations. Our evaluation includes, one-step and two-step question transformations, multi prompt testing and instruction guided assessment. We analyze a range of state-of-the-art clinical models and their general-purpose counterparts, focusing on Llama 3.1–based models. Additionally, we introduce Marmoka, a family of lightweight 8B-parameter clinical LLMs for English and Spanish, developed via continual domain-adaptive pretraining on medical corpora and instructions. \\

\noindent RESULTS: The experiments show that clinical LLMs do not consistently outperform their general purpose counterparts on English clinical tasks, even under the proposed perturbation based benchmark. However, for the Spanish subsets the proposed Marmoka models obtain better results compared to Llama.\\

\noindent CONCLUSIONS: Our results show that, under current short-form MCQA benchmarks, clinical LLMs offer only marginal and unstable improvements over general-purpose models in English, suggesting that existing evaluation frameworks may be insufficient to capture genuine medical expertise. We further find that both general and clinical models exhibit substantial limitations in instruction following and strict output formatting. Finally, we demonstrate that robust medical LLMs can be successfully developed for low-resource languages such as Spanish, as evidenced by the Marmoka models.
\end{abstract}

\begin{graphicalabstract}
\includegraphics[width=\textwidth]{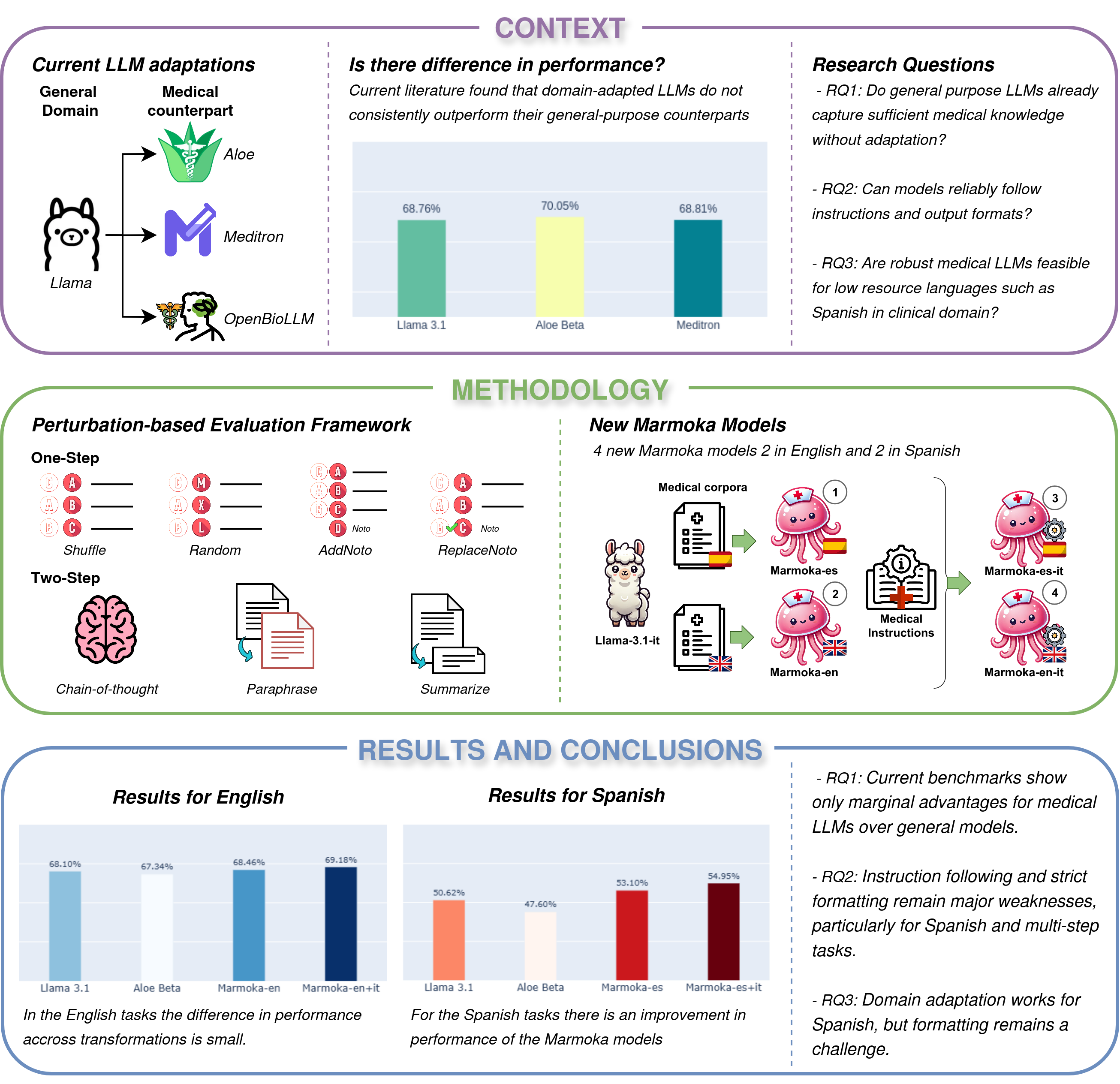}
\end{graphicalabstract}

\begin{highlights} 
\item Recent works found that domain adapted models do not outperform general counterparts.
\item The performance of general and clinical models is compared on different MCQA tasks.
\item A perturbation-based LLM evaluation benchmark is presented.
\item Four clinical LLMs are presented two for English and two For Spanish.
\item Clinical LLMs show no consistent gains in English, but improve in Spanish tasks.
\end{highlights}

\begin{keywords} 
Medical Large Language Models; Domain Adaptation; Perturbation based evaluation; Evaluation Frameworks; Clinical Question Answering;
\end{keywords}

\maketitle

\section{Introduction} \label{introduction}

Advances in large language models (LLMs) have enabled impressive performance across a variety of domains, including medical applications such as clinical note summarization, diagnostic assistance, and patient–provider interaction simulation. In response to this potential, many general-purpose LLMs have been adapted for the medical domain.

Recent works \citep{jeong2024medical,dorfner2024biomedicaLLMs,chen2025clinicalbench} found that domain-adapted LLMs do not consistently outperform their general-purpose counterparts. General-purpose models exhibit comparable or superior performance, occasionally, suggesting that these models may already possess substantial latent medical knowledge, raising important questions about the need for further domain specialization.

Nonetheless, caution is warranted. Emerging evidence \citep{alzahrani2024benchmarks,wiegreffe2024answer} shows that evaluation benchmarks are highly sensitive to even minor perturbations, which can lead to misleading conclusions about a model’s underlying capabilities. Moreover, the current predominant evaluation paradigm focuses on multiple-choice question-answering (MCQA) tasks, perplexity-based evaluation and accuracy metrics that do not fully capture real medical knowledge and the instruction following capacity of the models \citep{medicalLLMEvaluation,kweon2024ehrnoteqa,liu2025application}.

In addition, as LLMs evolve from research prototypes into practical clinical tools, there is a growing demand for smaller models (e.g., 7B–8B parameters) that can be realistically integrated into healthcare systems. Equally important is the development of models trained in minority languages beyond English, enabling their application across diverse health contexts.

The combined need for evaluation frameworks that extend beyond surface-level performance, together with the demand for lightweight, clinically usable models in minority languages, motivates the following research questions:
\begin{itemize}
    \item \textbf{RQ1.} Do we need to invest substantial resources in adapting general-purpose LLMs to the medical domain, or do these models already encode sufficient medical knowledge?
    \item \textbf{RQ2.} Are general-purpose and medical-domain models able to follow instructions and adhere to strict output formats?
    \item \textbf{RQ3.} Can robust medical-domain LLMs be developed for languages like Spanish, which remain low-resource in the medical domain due to the limited availability of high-quality medical data and instructions?
\end{itemize}
\vspace{-0.2cm}

To address the questions outlined above, this work introduces a comprehensive medical benchmark designed to evaluate LLMs in multiple MCQA tasks for both English and Spanish. Our framework includes multi-prompt evaluation, adversarial testing, and instruction-guided assessment to more rigorously examine the depth of a model’s medical understanding.

In response to the last question, we publicly release Marmoka\footnote{All training details, models, and the full training recipe are publicly available at \url{upponAcceptance}}, a family of Llama 3.1-Instruct models adapted for the medical domain in both English and Spanish. Focusing on the 8B variants to meet realistic healthcare needs, Marmoka is trained with a hybrid strategy that combines Domain Adaptation Pretraining (DAPT) and general instruction fine-tuning \citep{sainz2025instructing}, ensuring robust domain adaptation while preserving instruction-following ability.

\section{Related Work} \label{related_work}

The development and deployment of LLMs in the medical domain pose several key challenges, including domain-specific adaptation, adaptation strategy, model size, multilingual support and output format-following ability.


There is an ongoing debate regarding the actual \textbf{benefit of developing domain-specific medical LLMs}. Several studies have demonstrated that general-purpose LLMs can perform comparably to specialized biomedical models in clinical reasoning and knowledge tasks \citep{dorfner2024biomedicaLLMs, chen2025clinicalbench}. This raises critical questions about the cost-effectiveness and long-term value of investing in domain-specific adaptations when generalist models already exhibit strong medical capabilities.

In particular, \citet{jeong2024medical} provide a critical evaluation of medical LLMs, conducting a head-to-head comparison between each adapted model and its general-purpose base counterpart. Contrary to many previous claims, their findings indicate that open-source medical LLMs often achieve only marginal improvements across a variety of medical QA tasks. While the approach of directly comparing specialized models to their general-purpose source model is methodologically appropriate, it raises the question of whether assessing performance solely through simple MCQA tasks is sufficiently robust to conclude that domain-specific LLMs do not offer meaningful advantages in medical applications. 

In this work, we aim to investigate this limitation by proposing a more perturbation-based evaluation framework. This hypothesis is based on work that showed that LLMs can be brittle under small perturbations \citep{alzahrani-etal-2024-benchmarks}. Consequently, recent work has scaled up these ideas, proposing perturbations for MCQA tasks so that correct answers cannot be retrieved by memorization \citep{salido2025none, alzahrani-etal-2024-benchmarks, salemAdv}.


Regarding the \textbf{domain adaptation strategy}, we introduce the Marmoka models in the context of the recent growth of open medical LLMs, which are mainly developed through DAPT or instruction fine tuning of general domain models \citep{WANG2025103268}. 

Representative examples of DAPT include Meditron \citep{meditron}, based on Llama 2 \citep{llama2} and further pretrained on PubMed and clinical guidelines, and BioMistral \citep{biomistral}, which adapts Mistral using PubMed Central data. 

Instruction fine tuned approaches include Clinical Camel \citep{clinicalcamel}, built on Llama 2 with medical dialogues, and OpenBioLLM \citep{OpenBioLLMs}, which extends Llama 3 using direct preference optimization and a diverse medical instruction dataset. More recently, Aloe models \citep{aloe} built on Llama 3.1 and Qwen2.5 \citep{qwen2025qwen25technicalreport} use large scale supervised instruction tuning and model merging across multiple medical tasks, and despite not yet being peer reviewed, they can be considered the current state of the art in medical LLMs.

However, to the best of our knowledge, no prior work has explored the hybrid training strategy that combines DAPT with general instruction fine-tuning, as proposed in \cite{sainz2025instructing} and adopted for the Marmoka models.


The issue of \textbf{model size} remains a prominent topic of discussion. Although larger models often demonstrate better performance compared to their smaller counterparts \cite{dorfner2024biomedicaLLMs}, this advantage is not universally consistent. Recent studies comparing models of different scales within the same series have shown that an increase in size does not necessarily guarantee better performance in clinical tasks \citep{chen2025clinicalbench, domingo2026leveraging, DELBARRIO2026103379}. Furthermore, as highlighted in \citet{algaradi2025llmshealthcare}, there is an urgent need to prioritize resource-efficient architectures specifically optimized for medical applications, in contrast to the prevailing trend of constantly increasing model size. 


Another significant challenge is the development of \textbf{non-English medical LLMs}. The scarcity of large-scale, high-quality clinical datasets in languages other than English severely limits the creation and evaluation of multilingual models \citep{zheng2024efficiently, garcia-2024-medmt5}. This linguistic imbalance hinders the equitable deployment of AI-driven healthcare solutions worldwide. In response to this gap, the present work introduces a medical LLM specifically trained for Spanish-language clinical tasks.


In addition, modern LLMs generally excel at producing fluent text; however, many real-world applications require \textbf{strictly formatted outputs} (e.g., text classification, multiple-choice QA) \citep{liu2024llms}. Existing evaluation benchmarks, such as \texttt{lm-evaluation-harness} \citep{eval-harness}, primarily rely on selecting the lowest-perplexity option. As a result, they fail to assess a model’s ability to generate outputs in the desired format under open-ended generation settings.

In summary, while the necessity of domain-specific adaptation remains a subject of debate, simultaneously, the community faces critical hurdles in scaling models effectively, bridging the linguistic gap for non-English healthcare contexts, and ensuring models can adhere to the rigid output formats required for clinical integration. In the following sections, we will discuss these core themes in depth, evaluating how our proposed framework and the Marmoka models address the challenges of adaptation, multilingualism, and robust performance evaluation.

\section{Experimental setup}

This study presents a rigorous evaluation of a widely adopted small, open-source, general-purpose LLMs (Llama 3-Instruct, Llama 3.1-Instruct and Mistral) and their various of their domain-specific adaptations. Subsequently, focusing on Llama 3.1 variants enables rigorous and controlled comparisons, isolating the effects of domain specialization while minimizing architectural and pretraining differences.

Building upon the work of \citet{jeong2024medical} and the research questions presented in \autoref{introduction}, we extend the evaluation in three key methodological directions: (i) an analysis of the format output adherence, (ii) enhancing the evaluation framework by incorporating a range of controlled perturbations to assess model robustness, and (iii) expanding the linguistic scope to include Spanish.

\subsection{Evaluation Datasets}

The evaluation of the proposed models is conducted using eight state-of-the-art clinical datasets, comprising five in English and two in Spanish. A detailed description of the datasets, including their characteristics and sources, is provided in \autoref{tab:datasets}.

\begin{table*}[htb!]
\footnotesize
\centering
\begin{adjustbox}{max width=\linewidth}
\begin{tabularx}{1.5\textwidth}{lp{0.2\linewidth}cccX}
\toprule
\textbf{} & \textbf{} & \multicolumn{2}{c}{\textbf{\# Instances}} & \multicolumn{1}{l}{} & \textbf{} \\ \cmidrule(lr){3-4}
\textbf{Language} & \textbf{Dataset} & \textbf{Dev} & \textbf{Test} & \multicolumn{1}{l}{\textbf{\# Options}} & \textbf{Description} \\ \midrule

\multirow{5}{*}{\textit{EN}}    & MMLU \par \citep{mmlu}  & - & 1,089 & 4 & Massive multitask test consisting of multiple-choice questions from various branches of knowledge. A subset with the clinical tasks was extracted.\\
                                & PubmedQA \par \citep{pubmedqa} & 11.2k & - & 3 & Research biomedical questions with options yes/no/maybe using PubMed abstracts. \\
                                & MedQA \par \citep{medqa}  & 1.27k & 1.27k & 4 &  A multiple-choice adaptation for solving medical problems collected from professional medical board exams. \\
                                & MedMCQA \par \citep{medmcqa} & 4.18k & *6.15k & 4 &  MCQA dataset designed to address real-world medical entrance exam questions. This dataset is only used for model development assessment.\\
                                & CareQA (en) \par \citep{careqa}  & - & 5.62k & 4 & Healtcare MCQA dataset in English from official sources of the Spanish Specialized Healthcare Training examinations. \\
                                \\
                                \hdashline
                                \\
\multirow{3}{*}{\textit{ES}}    & Casimedicos-exp \par \citep{casimedicos}  & 63 & 125 & 4-5 & MCQA based of Resident Medical Intern exams in Spanish.\\
                                & CareQA (es) \par \citep{careqa} & - & 5.62k & 4 & Healtcare MCQA dataset in Spanish from official sources of the Spanish Specialized Healthcare Training examinations. \\
                                \bottomrule
\end{tabularx}
\end{adjustbox}
\caption{Summary of the datasets used to evaluate Marmoka models across various biomedical tasks.}
\label{tab:datasets}
\end{table*}

\begin{figure}[htb!]
    \centering
    \includegraphics[width=7cm]{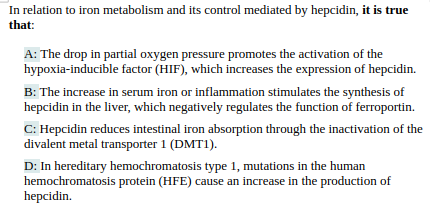}
    \caption{Instance of the CareQA-En dataset.}
    \label{fig:CareQA}
\end{figure}

\subsection{Evaluation Framework}

The proposed evaluation framework is designed to answer the research questions of \autoref{introduction}. The following sections describe each of the characteristics of our evaluation framework.

\subsubsection{Output Format Adherence}
In our customized implementation, the model generates a free-form answer that is directly compared to the correct answer, without any intermediate perplexity-based selection. Models are prompted to return only a single letter (A, B, C, or D). If the model generates anything beyond that, such as an explanation or a punctuation mark, the answer is considered incorrect. In other words, we evaluate based on the generated output, rather than the model’s output logits.

This design aims to assess not only the model’s reasoning ability but also its capacity to accurately follow instructions. This requirement is applied across all one-step perturbations or adversarial transformations.

Note that in existing evaluation benchmark, for instance \texttt{lm-evaluation-harness}, the response is usually selected by computing the perplexity for each option and choosing the option with the lowest perplexity. This choice is then compared against the ground-truth answer for evaluation. This setup often yields better results than open-ended generation because it simplifies the task to selecting among fixed options.

\subsubsection{Perturbations and Task Decomposition}

Firstly, to assess not only the accuracy of models in solving medical tasks but also the robustness of their underlying medical knowledge, we introduce a comprehensive evaluation framework that incorporates diverse input perturbations and decomposes tasks into well-defined solution steps. Our framework evaluates the original tasks with no alterations as in \citet{jeong2024medical} but additionally, it integrates the following transformations:

\paragraph{\textbf{One-Step Transformations}} The models are prompted to answer questions based on the given possible answer options.    
\begin{itemize}
    \item \textbf{Shuffled options (Shuffle)}: The order of the multiple-choice options is randomly shuffled. This evaluates whether the model's responses rely on fixed position rather than understanding the content of each option. Shuffling is included in all the following perturbations and two-step approaches.
    \item \textbf{Randomized option labels (Random)}: The traditional alphabetical option labels (A, B, C, D) are replaced with randomly assigned, non-sequential letters (e.g., M, Q, F, Y).
    \item \textbf{Add ``None of the Others'' (AddNoto}): An additional option labeled ``None of the others'' is introduced alongside the original answer choices. 
    \item \textbf{Replace with ``None of the Others'' (ReplaceNoto)}: The correct answer is removed and replaced by ``None of the others'', which now becomes the correct choice. 
\end{itemize}

\paragraph{\textbf{Two-Step  Transformations}} These strategies requires the model to engage in intermediate reasoning steps before answering.
\begin{itemize}
    \item \textbf{Chain-of-thought (CoT)}: The model is prompted to first generate a step-by-step rationale or argumentation for the question and then, in a new instance that includes this argumentation, provide the final answer based on the question and the reasoning.
    \item \textbf{Summary (Summ)}: The original question is first summarized, and the model must then answer based only on the summary.
    \item \textbf{Paraphrase (Par)}: Both the question and its options are rephrased while preserving their original meaning.
\end{itemize}

\subsubsection{Framework implementation}

Our evaluation framework follows the structure of the \texttt{lm-evaluation-harness} to ensure experimental consistency and reproducibility, with adaptations to support open ended generation rather than perplexity based evaluation. Wherever possible, we reused existing dataset implementations to maintain methodological alignment, while \textit{Casimedicos-exp} was implemented manually following the same design principles.

In addition, each dataset is evaluated across three independent runs and three distinct prompt configurations, with model specific optimization of the chat template and system prompt.

\subsection{Evaluated Models}

\subsubsection{Marmoka Models}

Based on \citet{sainz2025instructing}, we apply a modified DAPT approach to train our Marmoka medical models. Unlike traditional setups, we use an instruction-tuned model as the source rather than a base model. Specifically, we choose Llama 3.1–8B-Instruct as the starting point.

We train\footnote{\url{https://github.com/axolotl-ai-cloud/axolotl}} both English and Spanish versions of the model using an updated version of the medical corpora in \citet{de2025eriberta}, gathered from various sources (see \autoref{table:training_data}). In total, we use 26B words for English and 1B words for Spanish; all the corpora have been processed and deduplicated using standard approaches\footnote{\url{https://github.com/Maits27/Deduplication}}. To avoid catastrophic forgetting, we also include general-domain instructions generated by Llama 3.1–70B-Instruct \footnote{\url{https://huggingface.co/datasets/HiTZ/Magpie-Llama-3.1-70B-Instruct-Filtered}} using Magpie \citep{xu2024magpiealignmentdatasynthesis}: 1.4M for English and 310K for Spanish. This ensures both the inclusion of medical knowledge and the model’s ability to follow instructions. Note that this setup includes medical corpora, but not medical instructions\footnote{The amount of medical instructions generated by Magpie is relatively minimal.}.

\begin{table}[htb!] 
    \centering
    \begin{adjustbox}{max width=\linewidth}
\begin{tabular}{lll}
\toprule
\textbf{Language}   & \textbf{Dataset} & \textbf{Number of words} \\ \midrule
\multirow{3}{*}{English} & WikiMed        & 54.4M                     \\
                    & PubMed \cite{PubMedCorpus}           & 25.9B                     \\
                    & EMEA \cite{emea_source}             & 12.9M                     \\ \midrule
\multicolumn{2}{l}{\textbf{Total}}     & \textbf{26B}              \\ \midrule
\multirow{5}{*}{Spanish} & WikiMed        & 16.3M                     \\
                    & PubMed           & 5.5M                      \\
                    & MedCrawler \cite{medical_crawler}       & 850.7M                    \\
                    & MeSpEn \cite{mespen}            & 6.9M                      \\
                    & SciELO \cite{scielo}           & 29M                       \\ \midrule
\multicolumn{2}{l}{\textbf{Total}}     & \textbf{908.3M}           \\ \bottomrule
\end{tabular}
\end{adjustbox}
\caption{Word counts of medical raw text used for model training.}
\label{table:training_data} 

\end{table}

\begin{itemize}
    \item \textbf{Marmoka-en}: We train four English models using different hyperparameters, each on 6.5B unique words. These models are then merged into a single model using the MergeKit\footnote{\url{https://github.com/arcee-ai/mergekit}} tool \citep{goddard-etal-2024-arcees}.
    
    \item \textbf{Marmoka-es}: We train three Spanish models, each with different hyperparameters but using the same 1B word corpus. These are also merged into a single model.
\end{itemize}

For a fair comparison, we also use medical instructions to generate an additional model.

\begin{itemize}
    \item \textbf{Marmoka-it}: which includes 410K medical instructions from the UltraMedical\footnote{\url{https://huggingface.co/datasets/TsinghuaC3I/UltraMedical}} dataset \citep{zhang2024ultramedical}. As with the others, we use Llama 3.1-8B–Instruct as the source model. We further include 82K general-domain instructions and merge three models trained with different hyperparameters.
\end{itemize}

\begin{figure}[htb!]
    \centering
    \includegraphics[width=\columnwidth]{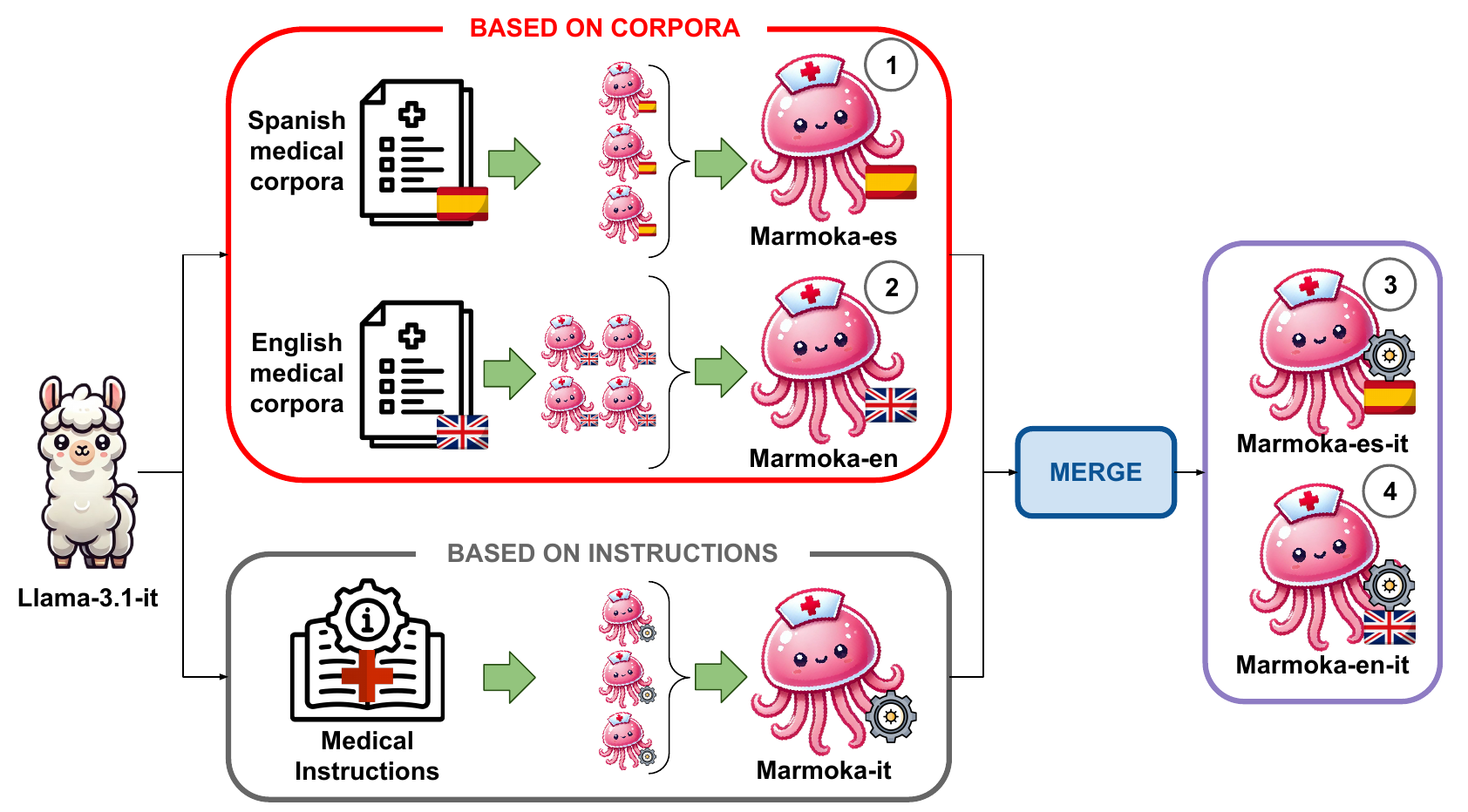}
    \caption{Overview of the Marmoka variants based on Llama 3.1-8B-Instruct. Four final models were developed: two trained on medical corpora (one in Spanish and the other in English), and two obtained by merging these models with an intermediate Marmoka model trained using medical instruction data.}
    \label{fig:marmoka-diagram}
\end{figure}

\autoref{fig:marmoka-diagram} shows the graphical abstract of the models presented above. Finally, we merged each of the \textit{Marmoka-en} and \textit{Marmoka-es} models with \textit{Marmoka-it} to combine the strengths of the different approaches. This resulted in two combined models: \textbf{Marmoka-en+it} and \textbf{Marmoka-es+it}.

In addition, to ensure robust evaluation and avoid benchmark contamination (a common issue in biomedical NLP where test data leaks into training) our methodology strictly separates training, development, and test sets. Following best practices, we use the development set exclusively for tuning and reserve the test set for final evaluation. For Marmoka models development, we employed only the Casimedicos \citep{casimedicos} validation set (for both English and Spanish) to select hyperparameters and guide model merging.

\subsubsection{Other Models}

The baseline models used for comparison with Marmoka are selected from \citet{jeong2024medical}, and we also include models from \citet{aloe} based in the Llama 3.1 architecture. All details can be found in \autoref{tab:models}.

\begin{table}[!hbt]
    \centering
    \begin{adjustbox}{max width=\linewidth}
    \begin{tabular}{l l l }
        \toprule
        \textbf{Model name} &  & \textbf{Architecture}\\ 
        \midrule
        Meditron-7b & \citep{meditron} & Llama 2 \\
        \midrule
        Llama-3.1-8B-Instruct & \citep{llama3} & Llama 3 \\
        Llama3-Aloe-8B-Alpha & \citep{aloe} & Llama 3 \\
        Llama3.1-Aloe-Beta-8B & \citep{aloe} & Llama 3 \\
        OpenBioLLM-8B & \citep{OpenBioLLMs} & Llama 3 \\ 
        Meditron3-8B & \citep{meditron3} & Llama 3 \\
        \midrule
        Mistral-7B-Instruct-v0.1 & \citep{mistral} & Mistral \\
        BioMistral-7B & \citep{biomistral} & Mistral \\
        \bottomrule
    \end{tabular}
    \end{adjustbox}
    \caption{Summary of the medical models evaluated against marmoka.}
    \label{tab:models}
\end{table}

\section{Results and Discussion}

This section begins by evaluating the models’ ability to comply with the required output format (see \autoref{subsec:format}). Subsequently, \autoref{subsec:english} presents an evaluation equivalent to that of \citet{jeong2024medical} for the English tasks using both the one-step and two-step transformations. Finally, \autoref{subsec:spanish} introduces the results in the Spanish tasks.

\subsection{Output Format Adherence} \label{subsec:format}

Before evaluating the models with the proposed perturbation benchmark, we first conducted experiments to analyze the output format adherence of the previously evaluated models. These results are particularly important for two reasons: (i) to determine whether the models can correctly follow instructions and return answers in the required format, and (ii) to ensure that the perturbation-based benchmark captures genuine clinical reasoning errors rather than format-related mistakes.

\begin{figure}[htb!]
    \includegraphics[width=\columnwidth]{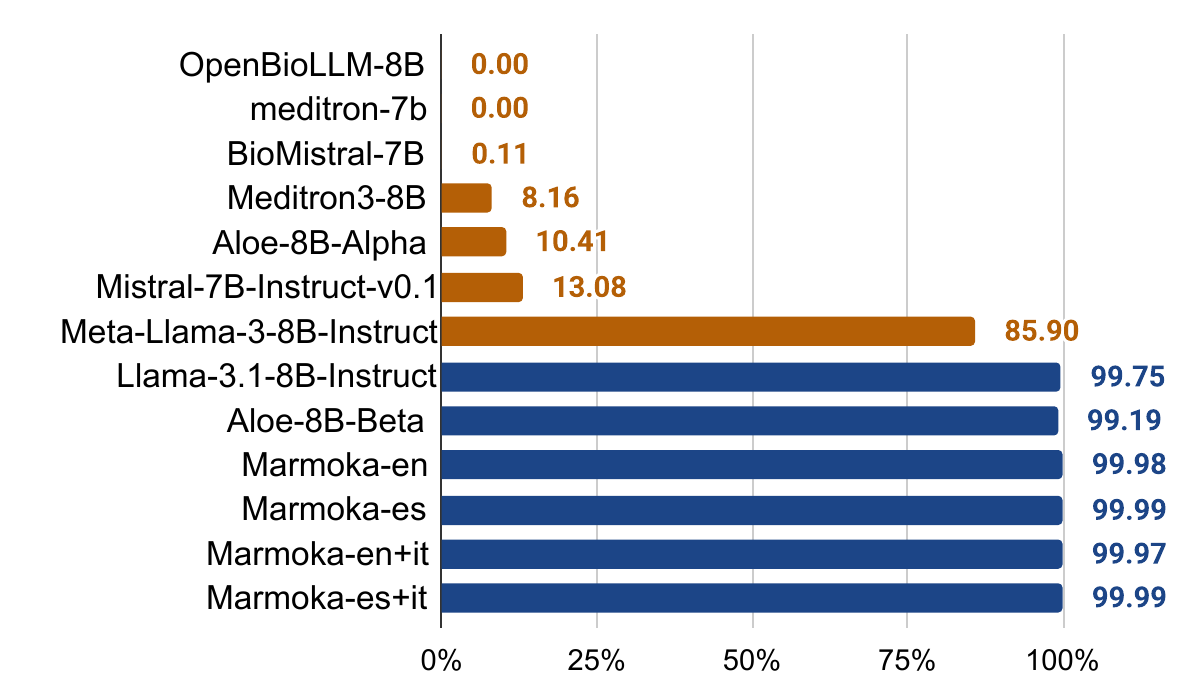}
    \caption{Percentage of answers correctly formatted according to task instructions for each evaluated model. Models shown in orange are excluded from further experiments due to a low rate of instruction adherence.}
    \label{fig:format-rate-plot}
\end{figure}

The results in \autoref{fig:format-rate-plot} reveal that many models fail to produce the required output format, which consists solely of the letter corresponding to the correct answer. Both the domain-specific and medically adapted Mistral models return correctly formatted outputs only in a small fraction of cases. In contrast, Llama 3 achieves a substantially higher rate of correctly formatted outputs (85.90\%), although this performance is still insufficient for reliable use. Moreover, all medical variants of Llama 3 perform markedly worse than the base model, with some failing to produce the desired format even once.

In contrast, Llama 3.1 and all of its medical counterparts are able to consistently produce the desired output format and are therefore included in the following section, where we apply the proposed benchmark.

\subsection{Results for English Datasets} \label{subsec:english}

This section presents the results of the English tasks under both the one-step and two-step transformations, comparing the performance of Llama 3.1 model with its medical counterparts on unperturbed tasks as well as under adversarial settings.

\subsubsection{One-step Transformations} \label{subsec:one-step-en}

\autoref{tab:one-step_en} provides a detailed view of model behavior in the one-step scenario and under evaluation perturbations.

\begin{table*}[!htb]
    \centering
    \tiny
    \begin{adjustbox}{max width=\linewidth}
    \begin{tabular}{lllrrrrr}\hline
    Language &Transformation &Dataset &Llama 3.1 &Aloe-Beta &Marmoka-en &Marmoka-en+it \\\hline
    
    \multirow{20}{*}{English} &\multirow{4}{*}{None} &MMLU &72.90 &\cellcolor[HTML]{ddf2e8}0.97 &\cellcolor[HTML]{f2faf6}0.39 &\cellcolor[HTML]{e9f7f0}0.62 \\
    & &PubMedQA &81.40 &\cellcolor[HTML]{fcfefd}0.10 &\cellcolor[HTML]{fdf9f8}-0.80 &\cellcolor[HTML]{fefdfd}-0.20 \\
    & &MedQA &60.42 &\cellcolor[HTML]{fdf9f9}-0.78 &\cellcolor[HTML]{dff2e9}0.92 &\cellcolor[HTML]{b9e3ce}1.99 \\
    & &CareQA-En &65.93 &\cellcolor[HTML]{cceadb}1.46 &\cellcolor[HTML]{c8e9d9}1.57 &\cellcolor[HTML]{b4e1cb}2.12 \\ \cmidrule{2-7}
    
    &\multirow{4}{*}{Shuffle} &MMLU &71.44 &\cellcolor[HTML]{f7fcfa}0.23 &\cellcolor[HTML]{dcf1e7}1.00 &\cellcolor[HTML]{c1e6d4}1.76 \\
    & &PubMedQA &81.63 &\cellcolor[HTML]{fdfefe}0.07 &\cellcolor[HTML]{fdf9f8}-0.80 &\cellcolor[HTML]{fefdfd}-0.26 \\
    & &MedQA &61.77 &\cellcolor[HTML]{fdf9f8}-0.80 &\cellcolor[HTML]{fefefe}-0.03 &\cellcolor[HTML]{ccebdc}1.45 \\
    & &CareQA-En &65.52 &\cellcolor[HTML]{d2ede0}1.28 &\cellcolor[HTML]{c7e9d8}1.59 &\cellcolor[HTML]{b3e1ca}2.14 \\ \cmidrule{2-7}
    
    &\multirow{4}{*}{Random} &MMLU &69.93 &\cellcolor[HTML]{fefbfb}-0.49 &\cellcolor[HTML]{fdf9f9}-0.77 &\cellcolor[HTML]{fefdfc}-0.27 \\
    & &PubMedQA &80.03 &\cellcolor[HTML]{fefbfb}-0.43 &\cellcolor[HTML]{fef9f9}-0.73 &\cellcolor[HTML]{ebf7f1}0.57 \\
    & &MedQA &58.80 &\cellcolor[HTML]{fefffe}0.05 &\cellcolor[HTML]{e1f3ea}0.86 &\cellcolor[HTML]{a8dcc3}2.45 \\
    & &CareQA-En &62.87 &\cellcolor[HTML]{c3e7d5}1.70 &\cellcolor[HTML]{e0f3ea}0.88 &\cellcolor[HTML]{c5e8d7}1.64 \\ \cmidrule{2-7}
    
    &\multirow{4}{*}{AddNoto} &MMLU &66.44 &\cellcolor[HTML]{bfe5d3}1.81 &\cellcolor[HTML]{c3e7d5}1.70 &\cellcolor[HTML]{99d6b8}2.88 \\
    & &PubMedQA &80.23 &0.00 &\cellcolor[HTML]{fefaf9}-0.70 &\cellcolor[HTML]{f7fcfa}0.24 \\
    & &MedQA &56.10 &\cellcolor[HTML]{97d5b6}2.95 &\cellcolor[HTML]{92d3b3}3.08 &\cellcolor[HTML]{61c091}4.44 \\
    & &CareQA-En &57.05 &\cellcolor[HTML]{57bb8a}4.72 &\cellcolor[HTML]{b9e3cf}1.97 &\cellcolor[HTML]{8ad0ad}3.31 \\ \cmidrule{2-7}
    
    &\multirow{4}{*}{ReplaceNoto} &MMLU &33.37 &\cellcolor[HTML]{ec9d96}-13.73 &\cellcolor[HTML]{f6d2cf}-6.19 &\cellcolor[HTML]{f4c8c4}-7.63 \\
    & &PubMedQA &64.07 &\cellcolor[HTML]{e67c73}-18.37 &\cellcolor[HTML]{f9e2e0}-4.04 &\cellcolor[HTML]{c9e9da}1.53 \\
    & &MedQA &27.47 &\cellcolor[HTML]{e7857d}-17.02 &\cellcolor[HTML]{f2beba}-9.04 &\cellcolor[HTML]{efaca6}-11.60 \\
    & &CareQA-En &34.75 &\cellcolor[HTML]{ea938c}-15.03 &\cellcolor[HTML]{f8dbd9}-4.94 &\cellcolor[HTML]{f6d0cd}-6.48 \\
    \hline
    \end{tabular}
    \end{adjustbox}
    \caption{Accuracy scores of the one-step perturbations in the English Tasks. Green and red cells indicate improvements or drops compared to Llama 3.1, with darker shades showing larger differences.}
    \label{tab:one-step_en}
\end{table*}

In the \textbf{base scenario} without perturbations, we observe relatively small performance differences between general-purpose and clinical models across all benchmarks. These differences are generally modest and dataset dependent, with Marmoka-en+it showing slightly better performance across most tasks. However, these improvements are too small to support a strong claim of a meaningful performance advantage.

Importantly, these relative differences remain stable under the \textbf{Shuffle} and \textbf{Random} perturbations. In most cases, the variation with respect to the base scenario is limited to approximately $\pm2$ accuracy points in the best cases, and often less. This indicates that Shuffle and Random transformations do not significantly alter the relative ranking of the models.

The \textbf{AddNoto} transformation introduces larger changes in absolute performance and reveals some model-specific behaviors. In this setting, clinical models, particularly Marmoka variants, occasionally benefit more than the general model. In contrast, the \textbf{ReplaceNoto} transformation has a strong negative impact on all models, with substantial drops in performance across all datasets. This confirms ReplaceNoto as a highly adversarial setting that significantly alters task difficulty and disrupts standard multiple-choice reasoning \citep{elhady2025wicked, salido2025othersgeneraltechniquedistinguish}.

It is also noteworthy the impact differences between the \textbf{AddNoto} and \textbf{ReplaceNoto} scenarios: in the former, Llama 3.1 is the most affected model, whereas in the latter, it is the least affected. A possible explanation is that Llama 3.1 may have a stronger tendency to predict “None of the others.” As a result, AddNoto harms its performance because the option is always incorrect, while ReplaceNoto improves it because the option is always correct.

\subsubsection{Two-step Transformations} \label{subsec:two-step-en}

The results obtained in the Two-step transformations for the English tasks are available in \autoref{tab:two-step_en}. To ensure a fair analysis of the results, it is also necessary to assess the quality of the intermediate outputs produced by each transformation. Specifically, for the chain of thought transformation, the model should generate reasoning in the requested format without explicitly stating the correct answer. For the summarization transformation, the output should be a faithful summary of the original question, without altering the answer options. Finally, for the paraphrasing transformation, the generated question should represent a genuine rephrasing of the original while preserving the content and the order of the answer choices.

To analyze these characteristics, we conducted a human evaluation of five examples per dataset for each model and each transformation, except for the MMLU dataset, where one question from each category was analyzed. Consequently, a total of 21 examples were analyzed per model and transformation.

\begin{table*}[htb!]
    \centering
    \tiny
    \begin{adjustbox}{max width=\linewidth}
    \begin{tabular}{lllrrrrr}\hline
    Language &Transformation &Dataset &Llama 3.1 &Aloe-Beta &Marmoka-en &Marmoka-en+it \\ \hline
    
    \multirow{16}{*}{English} &\multirow{4}{*}{None} &MMLU &72.90 &\cellcolor[HTML]{dff2e9}0.97 &\cellcolor[HTML]{f3faf6}0.39 &\cellcolor[HTML]{ebf7f1}0.62 \\
    & &PubMedQA &81.40 &\cellcolor[HTML]{fcfefd}0.10 &\cellcolor[HTML]{faeae8}-0.80 &\cellcolor[HTML]{fdf9f9}-0.20 \\
    & &MedQA &60.42 &\cellcolor[HTML]{fbeae9}-0.78 &\cellcolor[HTML]{e1f3ea}0.92 &\cellcolor[HTML]{bde4d1}1.99 \\
    & &CareQA-En &65.93 &\cellcolor[HTML]{ceecdd}1.46 &\cellcolor[HTML]{cbeadb}1.57 &\cellcolor[HTML]{b8e3ce}2.12 \\ \cmidrule{2-7}
    
    &\multirow{4}{*}{CoT} &MMLU &74.78 &\cellcolor[HTML]{f3fbf7}0.36 &\cellcolor[HTML]{fcf0f0}-0.54 &\cellcolor[HTML]{fcf1f0}-0.53 \\
    & &PubMedQA &81.07 &\cellcolor[HTML]{fdf8f8}-0.24 &\cellcolor[HTML]{f9e2e1}-1.07 &\cellcolor[HTML]{f9e0de}-1.17 \\
    & &MedQA &63.39 &\cellcolor[HTML]{daf0e5}1.12 &\cellcolor[HTML]{fae7e5}-0.91 &\cellcolor[HTML]{fafdfc}0.16 \\
    & &CareQA-En &69.61 &\cellcolor[HTML]{fdf5f4}-0.38 &\cellcolor[HTML]{fafdfc}0.16 &\cellcolor[HTML]{dff2e9}0.96 \\ \cmidrule{2-7}
    
    &\multirow{4}{*}{Summ} &MMLU &68.71 &\cellcolor[HTML]{e67c73}-45.21 &\cellcolor[HTML]{fefefe}-0.01 &\cellcolor[HTML]{ec9b95}-3.78 \\
    & &PubMedQA &80.20 &\cellcolor[HTML]{bde5d1}1.97 &\cellcolor[HTML]{f2faf6}0.40 &\cellcolor[HTML]{caeada}1.60 \\
    & &MedQA &61.90 &\cellcolor[HTML]{fdf6f5}-0.33 &\cellcolor[HTML]{eff9f4}0.49 &\cellcolor[HTML]{c9e9da}1.62 \\
    & &CareQA-En &69.12 &\cellcolor[HTML]{f9fdfb}0.19 &\cellcolor[HTML]{ecf7f2}0.59 &\cellcolor[HTML]{e4f4ec}0.83 \\ \cmidrule{2-7}
    
    &\multirow{4}{*}{Par} &MMLU &66.87 &\cellcolor[HTML]{fae5e3}-0.99 &\cellcolor[HTML]{f9e3e1}-1.07 &\cellcolor[HTML]{f8dcda}-1.31 \\
    & &PubMedQA &61.33 &\cellcolor[HTML]{57bb8a}10.10 &\cellcolor[HTML]{eff9f4}0.50 &\cellcolor[HTML]{5cbd8e}4.87 \\
    & &MedQA &58.60 &\cellcolor[HTML]{f9e4e2}-1.02 &\cellcolor[HTML]{d9f0e5}1.14 &\cellcolor[HTML]{c5e8d7}1.73 \\
    & &CareQA-En &58.70 &\cellcolor[HTML]{f0f9f5}0.47 &\cellcolor[HTML]{f3fbf7}0.36 &\cellcolor[HTML]{eff9f4}0.49 \\
        
    \hline
    \end{tabular}
    \end{adjustbox}
    \caption{Accuracy scores of the two-step perturbations in the English tasks. Green and red cells indicate improvements or drops compared to Llama 3.1, with darker shades showing larger differences.}
    \label{tab:two-step_en}
\end{table*}

The \textbf{CoT} transformation yields mixed effects. Although it improves performance on most datasets except PubMedQA, the relative differences between models remain small, with Llama 3.1 benefiting more from CoT prompting and obtaining better results than the clinical models. The human analysis of the intermediate step for the CoT transformation showed that all models consistently followed the required format across all datasets. The generated outputs included an analysis of each option, with up to three pieces of evidence in favor of and against each option. Moreover, the models respected the instruction not to explicitly state the correct answer at this stage

Regarding the \textbf{Summarization} transformation, it affects negatively all the models and the results indicate a high degree of sensitivity depending on the evaluation domain. The Aloe-Beta-8B model exhibits a critical vulnerability, suffering a substantial drop of -45.21 on MMLU compared to the Llama 3.1 baseline. Notably, the Marmoka-en+it model demonstrates a consistent advantage over Llama 3.1 in datasets such as MedQA, PubMedQA and CareQA-En, although there is a performance drop in the MMLU dataset. 

The performance degradation of Aloe-Beta is mainly due to inconsistent instruction following, particularly the failure to preserve answer options across datasets despite explicit constraints introduced in the prompt. Marmoka-en produces summaries that closely mirror those of Llama 3.1, resulting in minimal performance differences, whereas Marmoka-en+it generates more diverse summaries while retaining more relevant information. Interestingly, both Marmoka-en variants exhibit certain formatting failures on the MMLU dataset, including the removal of questions or answer options and the introduction of answer bias.

In contrast, the \textbf{Paraphrasing} transformation highlights a more uniform improvement across medical datasets for the adapted models. While all specialized models show a slight performance lag relative to Llama 3.1 on the MMLU dataset, they demonstrate significant gains in the others. The Aloe-Beta model achieves its most notable improvement in the PubMedQA dataset with a gain of +10.10, while the Marmoka-en+it model shows the most consistent robustness, maintaining positive differentials across all medical tasks including PubMedQA and MedQA. 

A manual revision of the paraphrasing outputs shows that Marmoka-en+it performs best, consistently preserving the original meaning, while Llama 3.1 and Marmoka-en also generate reliable paraphrases with acceptable stylistic changes and minimal impact on fidelity. In contrast, Aloe-Beta exhibits significant fidelity issues, introducing semantic alterations such as removing relevant clinical details or modifying demographic information, particularly in MedQA, which can change the correct answer. This behavior may be related to Aloe-Beta’s training on synthetically generated chain-of-thought data from MedQA, MMLU, and PubMedQA. In MedQA and MMLU, the chain-of-thought includes a initial summarization step that explicitly condenses the context, whereas in PubMedQA this summarization requirement is limited to the question. This difference may have encouraged Aloe-Beta to treat paraphrasing as a generative reformulation task rather than a fidelity-preserving transformation, increasing the likelihood of altering critical information.

\subsection{Results for Spanish Datasets} \label{subsec:spanish}

To further examine the performance of general and clinical models across different scenarios, this section presents the results for the Spanish tasks, enabling an analysis of model behavior in a minority language compared to English. As with the English experiments, the results are organized into one-step (See \autoref{subsec:one-step-es}) and two-step (See \autoref{subsec:two-step-es}) transformations.

\subsubsection{One-Step Transformations} \label{subsec:one-step-es}

\autoref{tab:one-step_es} shows the results for the Spanish tasks in the One-Step transformation scenario.

\begin{table}[htb!]
    \centering
    \tiny
    \begin{adjustbox}{max width=\linewidth}
    \begin{tabular}{lllrrrrr}\hline
    Language &Transformation &Dataset &Llama 3.1 &Aloe-Beta &Marmoka-es &Marmoka-es+it \\ \hline
    
    \multirow{10}{*}{Spanish} &\multirow{2}{*}{None} &Casimedicos-exp &50.80 &\cellcolor[HTML]{fcf2f1}-2.00 &\cellcolor[HTML]{aedfc7}4.40 &\cellcolor[HTML]{aedfc7}4.40 \\
    & &CareQA-Es &58.58 &\cellcolor[HTML]{f5fbf8}0.55 &\cellcolor[HTML]{d5eee2}2.29 &\cellcolor[HTML]{a8dcc3}4.70 \\ \cmidrule{2-7}
    
    &\multirow{2}{*}{Shuffle} &Casimedicos-exp &50.27 &\cellcolor[HTML]{fdf6f5}-1.34 &\cellcolor[HTML]{bfe6d3}3.46 &\cellcolor[HTML]{95d5b6}5.73 \\
    & &CareQA-Es &57.98 &\cellcolor[HTML]{eff9f4}0.89 &\cellcolor[HTML]{cbeadb}2.86 &\cellcolor[HTML]{a5dbc1}4.86 \\ \cmidrule{2-7}
    
    &\multirow{2}{*}{Random} &Casimedicos-exp &47.60 &\cellcolor[HTML]{fefdfd}-0.27 &\cellcolor[HTML]{bfe5d3}3.47 &\cellcolor[HTML]{8bd0af}6.27 \\
    & &CareQA-Es &54.67 &\cellcolor[HTML]{e2f3eb}1.61 &\cellcolor[HTML]{dcf1e6}1.94 &\cellcolor[HTML]{b0dfc8}4.28 \\ \cmidrule{2-7}
    
    &\multirow{2}{*}{AddNoto} &Casimedicos-exp &43.33 &\cellcolor[HTML]{a4dac0}4.94 &\cellcolor[HTML]{7fcba6}6.94 &\cellcolor[HTML]{57bb8a}9.07 \\
    & &CareQA-Es &50.30 &\cellcolor[HTML]{9ed8bc}5.25 &\cellcolor[HTML]{b8e3ce}3.84 &\cellcolor[HTML]{84cda9}6.68 \\ \cmidrule{2-7}
    
    &\multirow{2}{*}{ReplaceNoto} &Casimedicos-exp &24.00 &\cellcolor[HTML]{e67c73}-20.27 &\cellcolor[HTML]{fefcfc}-0.40 &\cellcolor[HTML]{fae5e3}-4.00 \\
    & &CareQA-Es &30.99 &\cellcolor[HTML]{e88981}-18.21 &\cellcolor[HTML]{f8dad8}-5.64 &\cellcolor[HTML]{f8dbd8}-5.51 \\
    \hline
    \end{tabular}
    \end{adjustbox}
    \caption{Accuracy scores of the one-step perturbations in the Spanish tasks. Green and red cells indicate improvements or drops compared to Llama 3.1, with darker shades showing larger differences.}
    \label{tab:one-step_es}
\end{table}

In the \textbf{base scenario}, Marmoka models consistently outperform Llama 3.1 across both Spanish datasets, confirming the benefits of domain adaptation for minority languages even in the absence of perturbations. Among them, the Marmoka-es+it variant generally achieves the strongest gains. Similarly, under the \textbf{Shuffle} and \textbf{Random} transformations, Marmoka models remain the most robust, showing smaller performance drops than their general domain counterparts.

As observed in the English experiments, the \textbf{AddNoto} transformation affects Llama 3.1 more severely, while clinical models remain comparatively stable, particularly the Marmoka variants. In contrast, the \textbf{ReplaceNoto} transformation has a stronger negative impact on clinical models, with Aloe-Beta experiencing substantial performance drops of up to 18 to 20 points, while Marmoka not being so affected. This behavior mirrors the English results, where \textbf{ReplaceNoto} emerges as the most challenging adversarial setting for all models.

Overall, Aloe-Beta exhibits mixed performance, with marginal gains on CareQA-Es and slight degradation on Casimedicos-exp. The lower performance of both Aloe-Beta and Llama 3.1 is expected, as neither is specifically trained on Spanish data, further underscoring the importance of domain adaptation when developing medical LLMs for minority languages.

\subsubsection{Two-Step Transformations} \label{subsec:two-step-es}

The experiments regarding the two-step transformation for the Spanish tasks are available in \autoref{tab:two-step_es}. In the two-step transformation experiments on the Spanish dataset, we also manually evaluated the effect of the system prompt language at each step. 

\begin{table}[!htb]
    \centering
    \tiny
    \begin{adjustbox}{max width=\linewidth}
    \begin{tabular}{lllrrrrr}\hline
    Language &Transformation &Dataset &Llama 3.1 & Aloe-Beta &Marmoka-es &Marmoka-es+it \\ \hline
    
    \multirow{8}{*}{Spanish} &\multirow{2}{*}{None} &Casimedicos-exp &50.80 &\cellcolor[HTML]{fdf5f5}-2.00 &\cellcolor[HTML]{9ed8bc}4.40 &\cellcolor[HTML]{9ed8bc}4.40 \\
    & &CareQA-Es &58.58 &\cellcolor[HTML]{f3fbf7}0.55 &\cellcolor[HTML]{cdebdc}2.29 &\cellcolor[HTML]{98d5b7}4.70 \\ \cmidrule{2-7}
    
    &\multirow{2}{*}{CoT} &Casimedicos-exp &48.85 &\cellcolor[HTML]{fcf3f3}-2.39 &\cellcolor[HTML]{7acaa3}6.02 &\cellcolor[HTML]{8cd1af}5.22 \\
    & &CareQA-Es &58.05 &\cellcolor[HTML]{d0ecde}2.17 &\cellcolor[HTML]{bfe5d3}2.91 &\cellcolor[HTML]{a3dabf}4.18 \\ \cmidrule{2-7}
    
    &\multirow{2}{*}{Summ} &Casimedicos-exp &36.79 &\cellcolor[HTML]{f2bfba}-13.67 &\cellcolor[HTML]{c1e6d4}2.82 &\cellcolor[HTML]{fdf7f7}-1.53 \\
    & &CareQA-Es &53.07 &\cellcolor[HTML]{e67c73}-28.00 &\cellcolor[HTML]{f8dbd9}-7.56 &\cellcolor[HTML]{fae5e3}-5.47 \\ \cmidrule{2-7}
    
    &\multirow{2}{*}{Par} &Casimedicos-exp &49.21 &\cellcolor[HTML]{fae4e2}-5.60 &\cellcolor[HTML]{eef8f3}0.80 &\cellcolor[HTML]{57bb8a}7.60 \\
    & &CareQA-Es &49.19 &\cellcolor[HTML]{fbeae8}-4.48 &\cellcolor[HTML]{f5fbf8}0.48 &\cellcolor[HTML]{9ad7b9}4.58 \\
    \hline
    \end{tabular}
    \end{adjustbox}
    \caption{Accuracy scores of the two-step perturbations in the Spanish tasks. Green and red cells indicate improvements or drops compared to Llama 3.1, with darker shades showing larger differences.}
    \label{tab:two-step_es}
\end{table}

An examination of the results table indicates that the outcomes are largely comparable to those of the One-Step transformations. The Spanish Marmoka models maintained their initial performance advantage throughout the adversarial tests, with the notable exception of the summarization adversarial, where their performance declined relative to Llama 3.1.

An analysis of a five-example subset from each dataset reveals distinct behavioral patterns in the generated paraphrases and summaries. Within the summarization task, both Aloe-Beta and Marmoka-es+it frequently struggled with format adherence, often removing answers from the questions despite explicit instructions to retain them. This accounts for the performance decline observed in the Marmoka-es+it model during the summarization task. Additionally, regarding the summarization objective itself, Aloe-Beta and Marmoka-es demonstrated a tendency to produce outputs that were either longer than the original text or unchanged in length. In contrast, Llama 3.1 consistently adhered to the required output format and accurately executed the summarization instructions.

In the paraphrase transformation, the required output format was generally maintained, with the exception of two instances in Aloe-Beta and one in Llama 3.1 where the order of answer options was altered. Regarding the quality of the paraphrasing itself, Llama 3.1 demonstrated the strongest performance, followed closely by Marmoka-es+it. Conversely, Aloe-Beta struggled significantly with this task, frequently altering the original meaning of the questions rather than providing an accurate paraphrase. A similar limitation was observed in Marmoka-es, though the impact was less severe.

\section{Conclusions}

This work challenges current trends in the evaluation of medical LLMs, demonstrating that there is still significant potential and that future research should build on the path we have established. Regarding the proposed research questions:

\textbf{RQ1. Do we need to invest substantial resources in adapting general-purpose LLMs to the medical domain, or do these models already encode sufficient medical knowledge?}

The results for the English tasks indicate that in short-form MCQA scenarios, the performance improvements of medical LLMs over general-domain models, such as Llama, are marginal (as demonstrated by \citet{jeong2024medical}) even under perturbed conditions. This trend persists in two-step transformations, where models frequently encounter difficulties adhering to complex instructions and specified output formats. Nevertheless, the Marmoka-en+it model introduced in this study consistently achieves the strongest results among the English variants.

These findings raise questions regarding the overall utility of English domain adaptation given the narrow performance gap. However, it is essential to note that while these quantitative experiments are extensive, the evaluation is subject to several interrelated limitations. The reliance on short-form MCQA may not provide a sufficient benchmark for assessing deep medical expertise; tasks requiring complex reasoning or broader clinical knowledge may be necessary to reveal the true capabilities of specialized models. Furthermore, because the reported scores rely exclusively on automated accuracy metrics without expert clinical adjudication, the validity of the models' underlying rationales remains unverified.

In summary, this research does not dismiss the necessity of specialized medical LLMs, but rather highlights the potential inadequacy of current evaluation frameworks in accurately measuring genuine medical proficiency.

\textbf{RQ2. Are general-purpose and medical-domain models able to follow instructions and adhere to strict output formats?}

The findings presented in \autoref{subsec:format} and across both two-step transformations (\autoref{subsec:two-step-en} and \autoref{subsec:two-step-es}) reveal a significant performance deficit regarding instruction following and strict adherence to output formats. 

Specifically, when models were required to return only a single letter within open-generation scenarios, many failed to comply, highlighting a fundamental limitation in their ability to maintain constrained formatting. This observation is further reinforced by the human evaluation of the two-step transformations.

This conclusion underscores the critical importance of evaluating the instruction-following capabilities of LLMs following domain adaptation, as specialized fine-tuning can lead to "catastrophic forgetting" of the original alignment and instruction-following behaviors. Furthermore, the difficulties observed in the Spanish tasks emphasize the necessity of incorporating multilingual instruction sets during the domain and language adaptation phases. Neglecting these elements in non-English contexts may result in models that possess the requisite medical knowledge but lack the linguistic agility to process and output information according to specific user requirements.

\textbf{RQ3. Can robust medical-domain LLMs be developed for languages like Spanish, which remain low-resource in the medical domain due to the limited availability of high-quality medical data and instructions?}


The results achieved by the Marmoka models, particularly Marmoka-es+it, demonstrate that the domain adaptation strategy employed in this study successfully facilitates the development of a robust, medical-domain Spanish LLM. In this context, the Marmoka models consistently outperformed both Aloe-Beta and Llama 3.1, confirming the necessity of specialized domain and language adaptation for Spanish medical tasks. 

However, the two-step transformations reveal a persistent limitation regarding output format adherence, indicating an area that requires further investigation to enhance the practical usability of the generated models.

\section{Future Work}

In the future, we plan to extend the proposed benchmark by including additional tasks, such as named entity recognition (a critical area where LLMs currently underperform) and disease diagnosis tasks. Finally, we aim to train smaller and reasoner models to fully leverage the deployment potential of the proposed Marmoka models.

\section*{Limitations}

Our trained Marmoka models are intended for research purposes and are not approved for clinical use. They must not be used as medical devices or relied upon for diagnostic or therapeutic decision-making. Model outputs should not be interpreted as medical advice. The use of these models without rigorous validation poses significant risks, including misinformation, misclassifications, and the potential for harm if relied upon in medical decision-making. Thus, when employed in actual clinical scenarios, all results must be independently reviewed and validated by qualified healthcare professionals.  

The training data for the Marmoka models, sourced from the web and open datasets, may not adequately capture the full range of linguistic, demographic, or clinical variability found in real-world medical settings. Thus, it may introduce various biases, including selection bias, demographic and linguistic underrepresentation, and unexamined ethical risks. These factors can lead to skewed predictions, unequal model performance across populations, and the potential reinforcement of existing health disparities and societal biases.

We did not conduct further assessments to determine whether the datasets used contain biases or personally identifiable information, nor did we independently verify the anonymization measures applied. However, all data sources employed in this work were previously processed, publicly released, and made available with stated limitations, including steps taken to ensure data quality and privacy compliance.

Finally, the LLM selection was limited to 8B-parameter, non-reasoning models from the Llama 3.1 family. While this constraint reduces architectural diversity, it enables rigorous, controlled comparisons across the selected models. This setup allows us to isolate the impact of domain specialization and dataset characteristics, while minimizing the influence of architectural differences or variations in pretraining corpora.

\section*{Acknowledgments}

This work has been partially supported by the HiTZ Center and the Basque Government, Spain (Research group funding IT1570-22), as well as by MCIN/AEI/10.13039/5011 00011033 Spanish Ministry of Universities, Science and Innovation  by means of the projects:

EDHIA PID2022-136522OB-C22 (also supported by FEDER, UE), DeepR3 TED2021-130295B-C31 (also supported by European Union NextGeneration EU/PRTR).

I. de la Iglesia has been funded by the FPU grant of the Spanish Ministry of Science, Innovation and Universities (MCIU) (FPU23/03347).

A. G. Domingo-Aldama has been funded by the Predoctoral Training Program for Non-PhD Research Personnel grant of the Basque Government (PRE\_2024\_1\_0224).

M. Urruela has been funded by the Predoctoral Training Program for Non-PhD Research Personnel grant of the Basque Government (PRE\_2025\_1\_0177).

\section*{Ethics Statement}
This study does not involve human participants, patient intervention, or access to identifiable personal health information. All experiments were conducted using publicly available benchmark datasets and previously released language models. No private clinical records, protected health information, or sensitive personal data were collected, accessed, or processed as part of this research.

\bibliographystyle{cas-model2-names}
\bibliography{cas-refs}


\end{document}